\documentclass[10pt,twocolumn,letterpaper]{article}

\usepackage{cvpr}              

\usepackage{multirow}
\usepackage[marginal]{footmisC}
\usepackage{amsmath}
\usepackage{amssymb}
\usepackage{booktabs}
\usepackage{verbatim}
\usepackage{afterpage}
\usepackage{graphicx}

\usepackage[accsupp]{axessibility}

\usepackage[dvipsnames]{xcolor}

\definecolor{cvprblue}{rgb}{0.21,0.49,0.74}
\usepackage[pagebackref,breaklinks,colorlinks,linkcolor=red]{hyperref}

\def\paperID{8989} 
\def\confName{CVPR}
\def\confYear{2024}

\title{\raisebox{-0.2\height}{\includegraphics[height=1.5em]{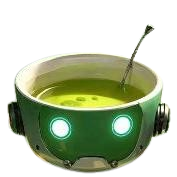}} MoCha-Stereo: Motif Channel Attention Network for Stereo Matching}
\author{
Ziyang Chen\textsuperscript{1}\thanks{Co-first author.}
\quad Wei Long\textsuperscript{1}\footnotemark[1] 
\quad He Yao\textsuperscript{1}\footnotemark[1]
\quad Yongjun Zhang\textsuperscript{1}\thanks{Corresponding author.}\\
\quad Bingshu Wang\textsuperscript{2}
\quad Yongbin Qin\textsuperscript{1}
\quad Jia Wu\textsuperscript{1}
\\
\textsuperscript{1}\footnotemark[2]~ College of Computer Science and Technology, The State Key Laboratory of Public Big Data,\\Institute of Artificial Intelligence, Guizhou University\\
\textsuperscript{2} College of Software, Northwest Polytechnical University
\\{\tt\small ziyangchen2000@gmail.com; zyj6667@126.com\footnotemark[2]}
}

\begin{document}
\maketitle   
\begin{abstract}
Learning-based stereo matching techniques have made significant progress. However, existing methods inevitably lose geometrical structure information during the feature channel generation process, resulting in edge detail mismatches. In this paper, the \textbf{Mo}tif \textbf{Cha}nnel Attention Stereo Matching Network (\textbf{MoCha-Stereo}) is designed to address this problem. We provide the Motif Channel Correlation Volume (MCCV) to determine more accurate edge matching costs. MCCV is achieved by projecting motif channels, which capture common geometric structures in feature channels, onto feature maps and cost volumes. In addition, edge variations in 
the reconstruction error map also affect details matching, we propose the Reconstruction Error Motif Penalty (REMP) module to further refine the full-resolution disparity estimation. REMP integrates the frequency information of typical channel features from the reconstruction error.  MoCha-Stereo ranks \textbf{1st} on the KITTI-2015 and KITTI-2012 Reflective leaderboards. Our structure also shows excellent performance in Multi-View Stereo. Code is avaliable at \href{https://github.com/ZYangChen/MoCha-Stereo}{MoCha-Stereo}.
\end{abstract}

\section{Introduction}
\label{sec:intro}

\quad Stereo matching remains a foundational challenge in computer vision, bearing significant relevance to autonomous driving, virtualization, rendering, and related sectors \cite{zbontar2015computing,chen2024feature}. 
The primary goal of the assignment is to establish a pixel-wise displacement map, or disparity, which can be used to identify the depth of the pixels in the scene. 
Edge performance of disparity maps is particularly vital in techniques requiring pixel-level rendering, such as virtual reality and augmented reality, where precise fitting between the scene model and image mapping is essential \cite{scharstein2002taxonomy}. This underscores the need for a close alignment between the edges of the disparity map and the original RGB image. 

Traditional stereo matching relies on global \cite{tra_gol}, semi-global \cite{SGM}, or local \cite{tra_local} grayscale relationships between left and right view pixels. These methods struggle to fully leverage scene-specific prior knowledge. Achieving optimal results often involves human observation, this tuning process can be resource-intensive in scenes with complex images \cite{zbontar2016stereo}.
With the advancement of deep learning, learning-based methods \cite{zbontar2015computing,tankovich2021hitnet,li2022practical} have generally achieved better results. 
A case in point is RAFT-Stereo \cite{lipson2021raft}, which introduced a coarse-to-fine scheme by computing All-Pairs Correlation (APC). 
To comprehensively learn channel features, GwcNet \cite{guo2019gwc} proposes a method of computing correlations by grouping left and right features along the channel dimension, called Group-Wise Correlation (GWC). 
IGEV-Stereo \cite{igev2023xu} introduces Combined Geometry Encoding Volume (CGEV) , a cost calculation method that combines GWC \cite{guo2019gwc} and APC \cite{lipson2021raft}, this cost calculation approach has achieved state-of-the-art results.
There is also a body of research focused on obtaining more accurate matching results through post-processing of disparities. They \cite{recon_error,stereodrnet,2023DLNR} utilize CNN structures directly applied to the additional error maps in the hope of achieving better results.

Learning-based methods have achieved impressive results. However, numerous channels experience loss of geometric details during the generation of feature channels. 
This phenomenon leads to a mismatch in the representation of object edges. 
Loss of geometric edges in the channel is a challenging problem because each block of the neural network performs non-linear transformations \cite{van1986nonlinear},
excessive nonlinearity can saturate activations in some channels, and insufficient nonlinearity leads to inadequate values. 
It is difficult for deep learning to directly recover geometric details. 
As shown in the middle picture of Fig. \ref{motivation}, certain normal channels suffer from severe blurring. 
Details loss in channels naturally complicates the matching of edges. 
To address the above problems, we propose \textbf{Mo}tif \textbf{Cha}nnel Attention Stereo Matching Network (MoCha-Stereo). The core idea of MoCha-Stereo is to restore the lost detailed features in feature channels by utilizing the repeated geometric contours within normal channels. 
Channels that preserve the common features are referred to as \textbf{motif channels}.
The following improvements are presented: 
\\1) We introduce a novel stereo matching framework that incorporates repeated geometric contours. 
This architecture enables more accurate cost computation and disparity estimation through detail restoration of feature channels.
\\ 2) We propose Motif Channel Attention (MCA) to mitigate imbalanced nonlinear transformations in network training. MCA optimize feature channels through motif channel projection instead of direct network optimization. Inspired by time-series motif mining, we capture motif channel using sliding windows. 
\\3) To achieve more precise matching cost computation for edge matching, we construct the Channel Affinity Matrix Profile (CAMP)-guided correlation volume. 
This volume is derived from the correlation matrix between normal and motif channels, then mapped onto the base correlation volume to produce a more rational cost volume called Motif Channel Correlation Volume (MCCV).
\\4) To leverage the geometric information of the potential channels in the reconstruction error map, we develop Reconstruction Error Motif Penalty 
(REMP) to extract the motif channels from the error map, optimising the disparity based on the high and low-frequency signals. 

\begin{figure}[]
	\centering
	\begin{subfigure}[b]{0.5\linewidth}
		\centering
		\includegraphics[width=\linewidth]{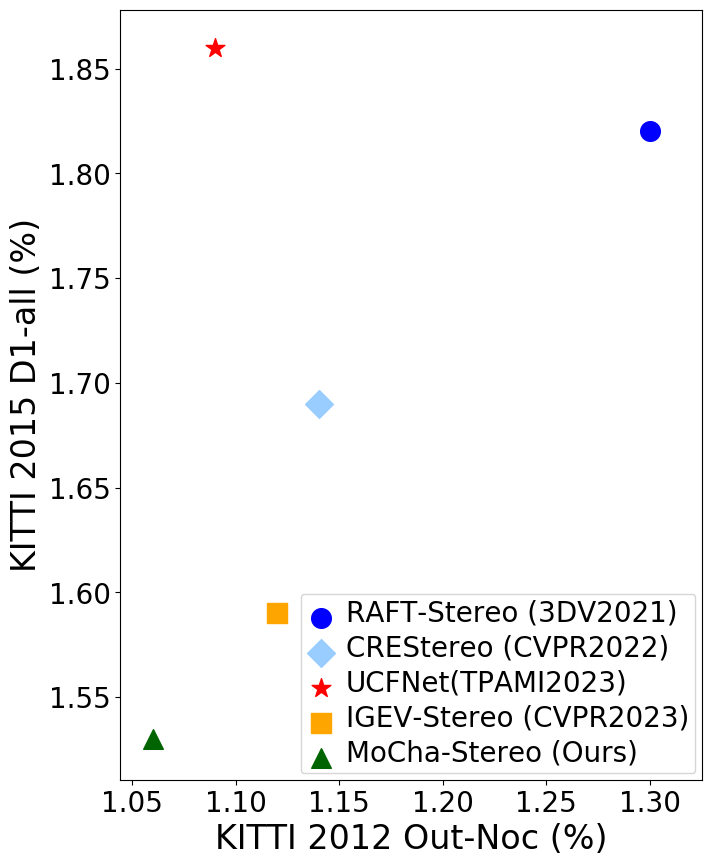}
		\caption{}
	\end{subfigure}
	\begin{subfigure}[b]{0.5\linewidth}
		\centering
		\includegraphics[width=\linewidth]{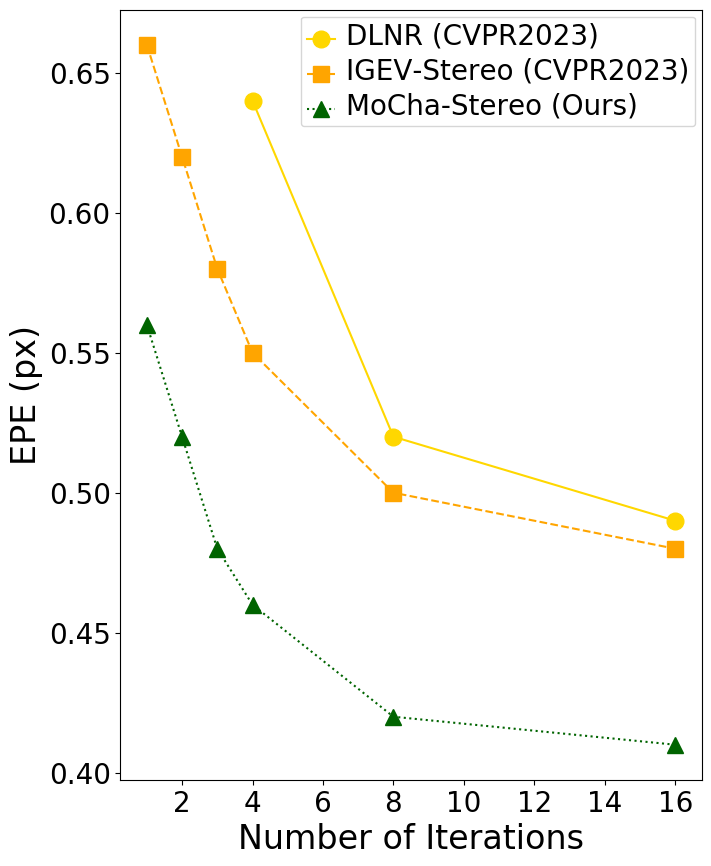}
		\caption{}
	\end{subfigure}
	\caption{(a) Comparison with SOTA methods \cite{lipson2021raft,ucfnet,li2022practical,igev2023xu} on KITTI 2012 \cite{kitti2012} and 2015 leaderboards \cite{kitti2015} (lower is better). (b) Performance evaluation of the Scene Flow test set \cite{sceneflow} in comparison to IGEV-Stereo \cite{igev2023xu} and DLNR \cite{2023DLNR} as the number of iterations changes (lower EPE means better). }
	\label{fig:2}
\end{figure}

We validated the performance of MoCha-Stereo on several leaderboards.
As shown in Fig. \ref{fig:2} (a), MoCha-Stereo ranks \textbf{1st} on KITTI 2015 \cite{kitti2015} and KITTI 2012 Reflective \cite{kitti2012}, achieves SOTA result in Scene Flow \cite{sceneflow}, zero-shot performance \cite{middlebury,eth3d}, and MVS domain \cite{jensen2014large}. 
Our designs also make iteration more efficient. As illustrated in Fig. \ref{fig:2} (b), MoCha-Stereo achieves superior results with fewer iterations, allowing users to choose between efficient or high-precision settings based on their preferences.

\begin{figure*}[]
	\centering
	\includegraphics[width=\linewidth]{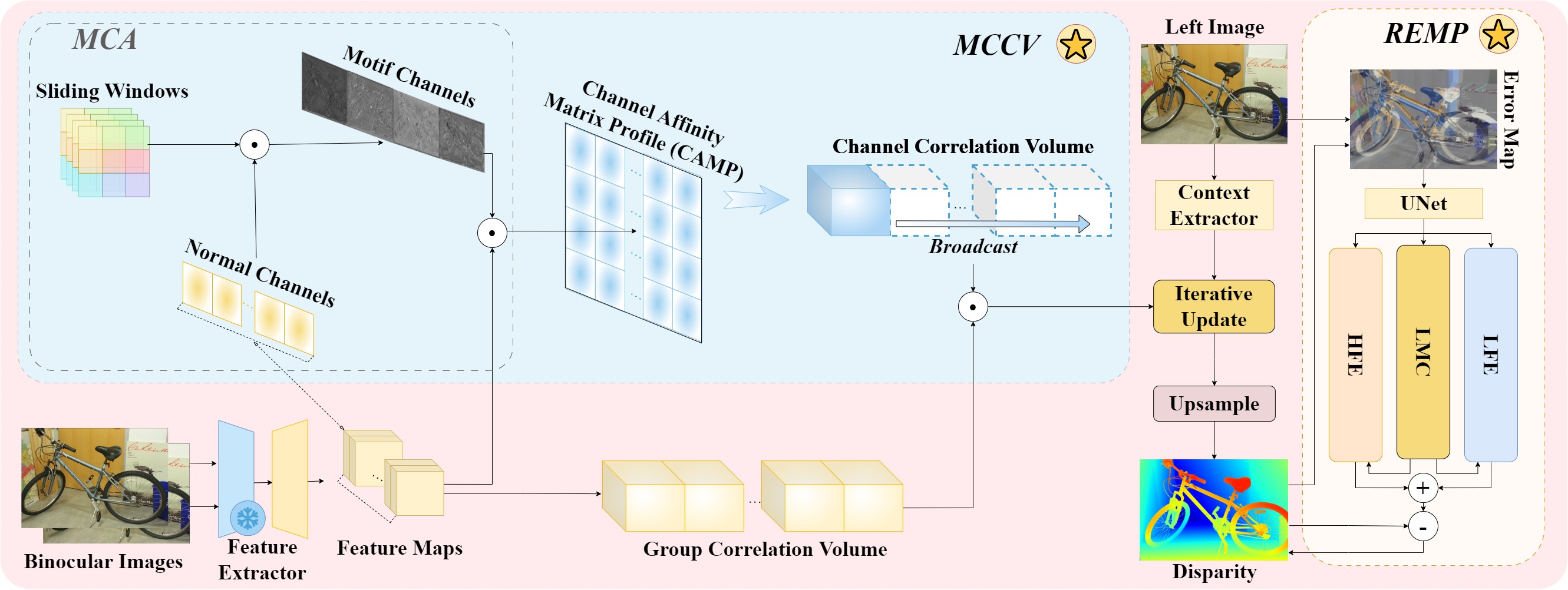}
	\caption{Overview of MoCha-Stereo. 
	MoCha-Stereo initially constructs the Motif Channel Correlation Volume (\textbf{MCCV}) by projecting the relationship between motif channels and normal channels into the basic group correlation volume. Subsequently, based on this cost volume, we employ an iterative way to build the disparity map. Finally, the Reconstruction Error Motif Penalty (\textbf{REMP}) module is applied to penalize the generation of the full-resolution disparity map. In REMP, $LFE$ refers to the Low-frequency Error branch, $LMC$ denotes to the Latent Motif Channel branch, and $HFE$ means the High-frequency Error branch.
	The \textbf{star symbol} means our primary innovations. 
	} \label{pipeline}
\end{figure*}
\begin {comment}
\begin{itemize}
	\item 
	Diverging from extant methodologies that depend on plane feature aggregation, we suggest employing View Frustum guidance for cost aggregation. 
	By aggregating features along the depth dimension, MoCha reduces the interference caused by ill-posed regions dominated by planar structures.	
	\item 
	We introduce the CNF designed to counteract the blurring of edges between the foreground and background, a prevalent issue in stereo matching tasks.
	\item 
	We suggest a semantic feature extractor tailored to learn external features, hence fortifying the constraint of semantic information. Specifically, We advocate for a dual-path feature extraction mechanism aimed at robust feature learning.
	\item 
	MoCha exhibits a potent cross-dataset generalization capacity. It consistently ranks \textbf{1st} on the Reflective KITTI-2012 and KITTI-2015 leaderboards. Furthermore, its architecture excels in achieving superior results in the MVS context.
\end{itemize}
\end {comment}

\section{Related Work}
\label{sec:formatting}
\subsection{Motif Mining for Time series analysis}
\quad The concept of motif originates from time series analysis. Motif mining has become one of the most commonly used primitives in time series data mining \cite{matrix1,matrix5}. In a time series $T$, there exists a subsequence $T_{i,L}$, which starts from the $i$-th position in the time series $T$ and is a continuous subset of values with a length of $L$. Motif is the pair of subsequences $T_{a,L}$ and $T_{b,L}$ in a time series that are the most similar. In mathematical notation, For case $\forall i,j \in [1,2,...,n-L+1]$ and $a \neq b\ , \ i \neq j$, the motif \cite{matrix22} satisfies as Equ. \ref{motif}.
\begin{gather}
	\label{motif}
	\begin{aligned}
		\centering
		dist(T_{a,L},T_{b,L}) &\leq dist(T_{i,L},T_{j,L}) 
	\end{aligned}
\end{gather}
where $dist$ means a distance measure. The distances between all subsequences and their nearest neighbors are stored as \textbf{M}atrix \textbf{P}rofile (\textbf{MP}) \cite{matrix1}, representing the repetitive pattern in the time series. 
This numerical relational information is utilized to identify recurring patterns, detect anomalies, and perform pattern matching in time series.

For the channel features, the geometric structure of graphics theoretically repeats as well. 
Differing from time series, the repetitive patterns we aim to uncover represent the geometric structures in images, necessitating consideration of two-dimensional contextual information. 
Furthermore, the computation of MP requires multiple samplings of subsequences from the series. Selecting sub-patches from multi-channel image features for computing similarity is computationally expensive.
Therefore, we develop a collection of adaptable sliding windows that are arranged into two-dimensional vectors. These windows are used for capturing repeating geometrical patterns. 
These repeated patterns are stored as motif channels for MoCha-Stereo.

\subsection{Learning-based Stereo Matching}
\quad The field of stereo matching has witnessed considerable advancements owing to learning-based techniques in recent years. 
Gwc-Net \cite{guo2019gwc} introduced the novel concept of GWC volume, which served as a major inspiration for future landmark achievements in the field \cite{pcw2022,Lac2022Liu}. 
RAFT-Stereo \cite{lipson2021raft} advances the construction of the cost volume by deploying an APC pyramid Volume. 
Building upon the foundation set by Gwc-Net \cite{guo2019gwc} and RAFT-Stereo \cite{lipson2021raft}, IGEV-Stereo \cite{igev2023xu} proposed a volume that combines APC \cite{lipson2021raft} and GWC \cite{guo2019gwc}, utilizing an iterative indexing method for the update of the disparity map. 
Another portion of learning-based methods are focuse on obtaining more accurate matching results through post-processing of disparities. iResNet \cite{recon_error} and DLNR \cite{2023DLNR} utilize UNet and Hourglass, respectively, employing convolutional operations to directly output the convolutional results of the reconstruction errors. DRNet \cite{stereodrnet} opts not to attach additional reconstruction error maps but instead appends Geometric and Photometric Error.

Learning-based methods have made significant progress. However, 
state-of-the-art algorithms \cite{igev2023xu,2023DLNR} inevitably lose geometric details in certain channels.
Few studies have considered using repeated geometric profiles in multiple feature maps to restore the edge texture of channel feature maps. Additionally, there is limited recognition regarding the significance of the frequency information in the potential motif channels from error maps for shaping the edges. 
\section{Method}

\quad To address the aforementioned issues, we respectively introduce MCCV and REMP 
as depicted in Fig. \ref{pipeline}. 
These components utilise the projection of Motif Channels to reinstate the geometric structure of channel features.


\subsection{Context Extractor \& Feature Extractor}
\quad Following \cite{lipson2021raft,igev2023xu}, the context extractor consisting of a series of residual blocks and downsampling layers, generating context at $1/4$, $1/8$, and $1/16$ scales.
The feature extractor, which also follows \cite{igev2023xu}, uses a backbone pretrained on ImageNet \cite{deng2009imagenet} as the frozen layer \cite{tan2019efficientnet}. The upsampling blocks utilizes skip connections from outputs of downsampling blocks to obtain multi-scale outputs ${f_{l,i}(f_{r,i}) \in \mathbb{R}^{C_i \times \frac{H}{i} \times \frac{H}{i}} }$ ( $i=4, 8, 16, 32$), $C_i$  represents the feature channels, and $f_{l}(f_{r})$ denotes the left (right) view features here.

\subsection{Motif Channel Correlation Volume}
\quad Although multi-channel extractors contribute to the learning of intricate features, an excess of nonlinearity may saturate activation values in specific channels, while insufficient nonlinearity can yield suboptimal activation values. 
MCCV is proposed to address the imbalanced learning of geometric structures in feature channels.\\
\textbf{Motif Channel Attention (MCA) for feature maps.} 
Feature channels exhibit varying degrees of geometric structure loss, but the fundamental geometric structure of feature channels is theoretically invariant. 
Inspired by motif mining for time series \cite{matrix5}, we use $N_{s}$ sets of sliding windows $SW$ ($N_{s}=4$), each with a length of $9$, and organized the windows into $3 \times 3$, to mine repetitive patterns. Unlike the window used for mining temporal motifs, we designed a set of adaptive-weight windows, rather than extracting values directly from the feature map. The initial values of the sliding window are set as random parameters. Based on the gradient loss, we backpropagate to adjust the values of the window weights. Following Equ. \ref{mcf}, we obtain the $s$-th ($1 \le s \le N_{s}$) motif channel feature map $f_{fre}^{mc}$ in the frequency domain.
\vspace{-10pt}
\begin{gather}
	f_{fre}^{mc}(s,h,w) =
	\sum_{c=1}^{N_c} \sum_{i=0}^{2} \sum_{j=0}^{2}	
	\nonumber	\\
	(SW(s,h+i,w+j) \times f_{fre}(c,h,w))
	\label{mcf}
	\vspace{-10pt}
\end{gather}
where $(h,w)$ are the coordinates of the pixel, $c$ denotes the $c$-th feature channel, $N_c$ is the number of normal feature channels. The frequency domain feature $f_{fre} = F(f-G(f))$. $F$ is the Fourier transform, 
$G$ denotes Gaussian low-pass filter with $3 \times 3$ kernels. 
This approach aims to capture repeatedly occurring frequency-domain features. The rationale behind this is that edge textures often exhibit high-frequency expressions.
We then transformed the results back to the spatial domain, accumulating and normalizing them to derive motif channels $f^{mc}$. 
This aggregation method takes into account the surrounding pixel information and strengthens attention to the geometric structure repeated across channels through accumulation, enhancing the reliability of matching for edge textures.\\
\textbf{Channel Affinity Matrix Profile (CAMP) guided Correlation Volume.} 
In order to enhance the accuracy of matching cost computation for edges with the assistance of motif channels, we propose the Correlation Volume guided by CAMP, as shown in Fig. \ref{pipeline}.
The foundational cost volume still employs the extraction of feature maps $f_{l(r),4}$ using GWC \cite{guo2019gwc} according to Equ. \ref{gwc}.
\begin{gather}
	\mathbb{C}_{g}(d,h,w,g) = \frac{1}{N_c/N_g}\langle f^g_{l,4}(h,w),f^g_{r,4}(h,w+d)\rangle
	\label{gwc}
\end{gather}
where $d$ is the disparity level,  $\langle \cdots,\cdots \rangle$ is the inner product, $N_c$ is the number of channels, $N_g$ is the group number ($N_g=8$).
Caculating Correlation Volume solely from feature maps makes it challenging to accurately match details. This difficulty arises because certain channels lose geometric structure. 
MoCha-Stereo exploits the motif channels $f^{mc}$ obtained by MCA and the normal channels $f_{l,4}$ obtained by feature extractor. Using their affinity, MoCha-Stereo constructs CAMP, which is a matrix that stores the relationship between motif channels and normal channels, as shown in Equ. \ref{CAMP}. 
\vspace{-10pt}
\begin{gather}
	CAMP(s,c,h,w) = f^{mc}(s,h,w) \times f_{l,4}(c,h,w), \nonumber
	\\
	where~~1 \leq s \leq N_s, 1 \leq c \leq N_c
	\label{CAMP}
\end{gather}
where $s$ denotes the $s$-th motif channel, and $c$ denotes the $c$-th normal channel. 
CAMP allows the projection of motif channels onto normal channels, serving as a coefficient to modulate the spatial domain information of normal channels. This enhances attention to the geometric structure of channels. 
We can construct a Channel Correlation ($\mathbb{C}_{c}$) based on CAMP. As shown in Equ. \ref{CC}, $\mathbb{C}_{c}$ performs spatial interaction across channels, theoretically reinforcing frequent patterns in space, which are the desired geometric structure.
\vspace{-10pt}
\begin{gather}
	\mathbb{C}_{c}(d,h,w) = \sum_{s=1}^{N_s} \sum_{c=1}^{N_c}
	\langle  3DConv(CAMP(s,c,h,w)), \nonumber
	\\3DConv(CAMP(s,c,h,w+d)) \rangle
	\label{CC}
\end{gather}
wher $3DConv$ means 3D convolution operator. 

To obtain the final cost volume $\mathbb{C}$, MoCha-Stereo uses $\mathbb{C}_{c}$ as a weight-adjusted basis for the basic Correlation Volume. Since geometric structures are theoretically invariant, there is no need to learn new $\mathbb{C}_{c}$ by adding extra groups. Broadcasting is sufficient to achieve the interaction between $\mathbb{C}_{c}$ and GWC $\mathbb{C}_{g}$, as illustrated in Equ. \ref{final}, enabling different groups to learn the same set of geometric structure features. 
\vspace{-15 pt}
\begin{equation}
	\mathbb{C}(d,h,w) = \sum_{g=1}^{N_g} (\mathbb{C}_{g}(d,h,w,g) \times \mathbb{C}_{c}(d,h,w) )
	\label{final}
\end{equation}

\subsection{Iterative Update Operator}
\quad Following \cite{zhao2022eai,2023DLNR,igev2023xu}, MoCha-Stereo utilizes the iterative update operator \cite{2023DLNR} to obtain the disparity map $d_{k} = d_{k-1} + \triangle d_k$ at 1/4 resolution for the $k$-th iteration.
\begin{figure}[]
	\centering
	\includegraphics[width=\linewidth]{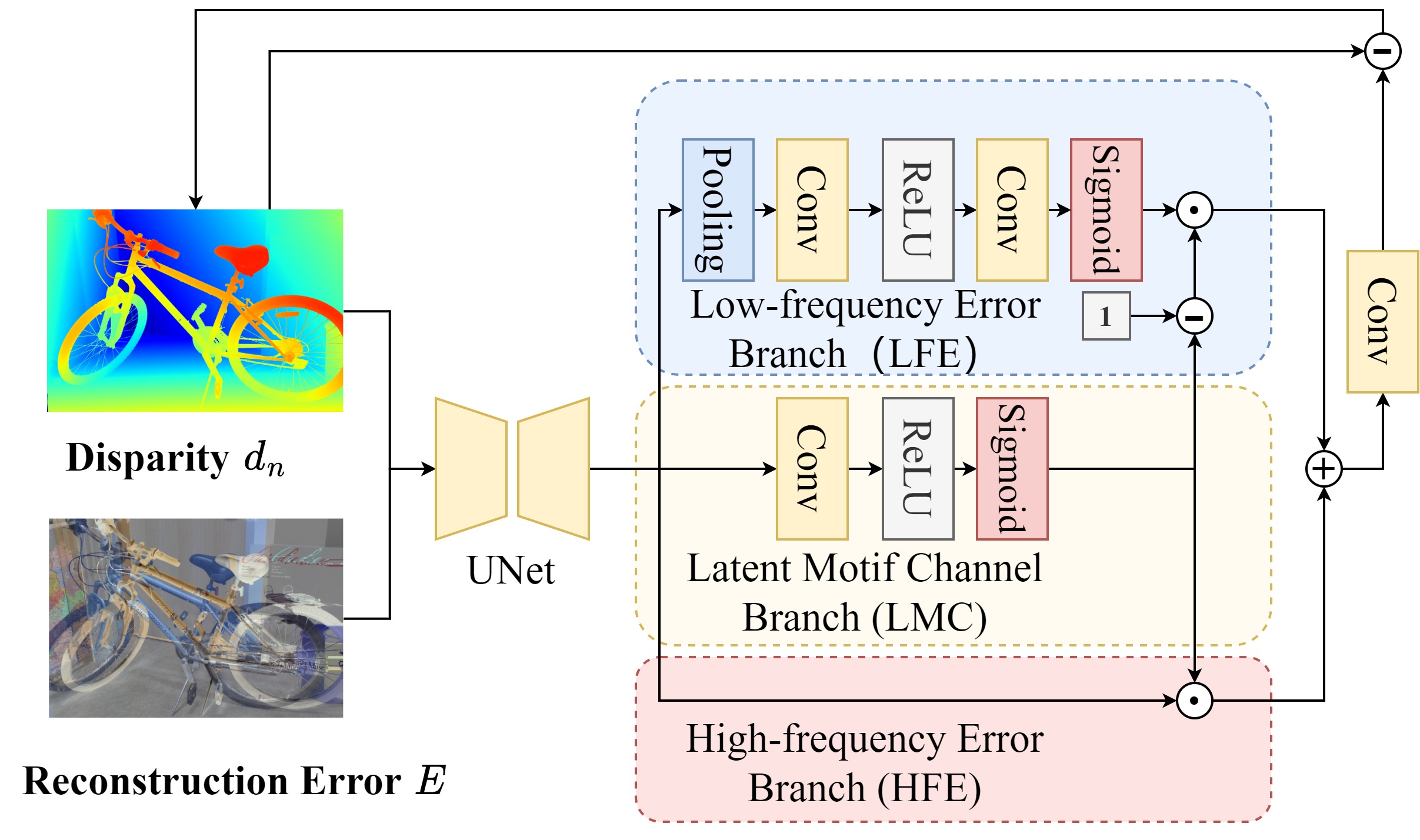}
	\caption{REMP module for Full-Resolution Refine. The upper branch (LFE) obtains low-frequency information through pooling, the lower branch (HFE) retains the original high-resolution image as high-frequency detailed information, and the middle branch (LMC) learns motif features through CNN.} \label{REMP}
\end{figure}
\subsection{Reconstruction Error Motif Penalty}
\quad The disparity map output by the iteration is at a resolution of 1/4 of the original image. There is still room for optimization after upsampling the disparity map.
Several works \cite{recon_error,zhao2022eai,2023DLNR} have been dedicated to refinement networks based on reconstruction error. However, none of these works have addressed the separation of the low-frequency and high-frequency components of the reconstruction error, making the refinement process challenging to achieve more effective results.
We propose the Reconstruction Error Motif Penalty (REMP) module for Full-Resolution Refine.
The disparity map is a single-channel image. 
To obtain multi-channel information, we draw inspiration from \cite{stereodrnet}, using the error as the input to the network and outputting multi-channel features. In contrast to \cite{stereodrnet}, we use reconstruction error $E$ obtained by Equ. \ref{warp} and the disparity map $d_n$ from the last iteration as inputs. 
\begin{gather}
	E = K_l (R-\frac{TN^T}{D}) K^{-1}_r I_{r} - I_l
	\label{warp}
\end{gather}
$K_{l(r)}$ represents the intrinsic matrix of the left (right) camera in the stereo system, $R$ is the rotation matrix from the right view coordinate system to the left view coordinate system, $T$ is the translation matrix from the right view coordinate system to the left view coordinate system, $N$ is the normal vector of the object plane in the right view coordinate system, $D$ is the perpendicular distance between the object plane and the camera light source (this distance is obtained from the computed disparity), $I_l(r)$ is the left (right) image.

As illustrated in Fig. \ref{REMP}, the UNet in REMP is solely designed to obtain multi-channel features related to the reconstruction error $E$ and disparity map $d_n$. The core idea of REMP is to optimize both high-frequency and low-frequency errors in the disparity map using representative motif information.
REMP divides the features output by the UNet into three branches. The pooling operation in the upper branch (LFE) effectively acts as a low-pass filter, preserving low-frequency information in the image while attenuating high-frequency information to some extent. The middle branch (LMC) guides the network in learning typical motif information, and the lower branch (HFE) undergoes no transformation, aiming to preserve the original high-frequency details of the high-resolution image. Through the mappings of these three branches, we learn the feature errors as penalties to refine the disparity map $d_n$, as shown in Equ. \ref{RE}.
\begin{gather}
	o = \mathit{UNet}(Concat(d_n^\prime ,E))	
	\nonumber \\
	HFE\left( o\right)  = o \odot LMC(o) \nonumber \\
	d_n = d_n^\prime - Conv(LFE(o) \odot (1-LMC(o)) + HFE(o))
	\label{RE}
\end{gather}
where $\odot$ means Hadamard product, $d_{n}^{\prime}$ represents the disparity map before refinement, $lfe$ denotes the computation process of low-frequency error, and $lmc$ refers to the computation in the LMC branch.
\subsection{Loss Function}
\quad The computation of the loss function requires the disparity maps outputted at each iteration as well as the initial disparity map $d_0$. 
The initial disparity $d_0$ is obtained from the Volume $V_g (g \in \{1,2,\ldots, N_g\})$ projected by $\mathbb{C}_{c}$ and GWC $\mathbb{C}_{g}$. 
For each group $g~(g \leq N_g)$, the cost calculation method $\mathbb{C}_g $ is uniquely associated with a corresponding volume $V_g$.
We generate $d_0$ through $N_g$ groups of correlation volumes, expressed by Equ. \ref{initdisp}.
\begin{equation}
	d_0 = SoftMax\left(3DConv\left(V_1 \oplus V_2 \oplus \ldots \oplus V_{N_g}\right)\right)	
	\label{initdisp}
\end{equation}
where $\oplus$ refers to the concatenation operation performed along the group dimension. The initial disparity $d_0$ serves as the starting point for the iteration and is input to the update module.
Following  \cite{igev2023xu}, the total loss is defined as Equ. \ref{losst}.
\begin{equation}
	L = \mathit{Smooth}_{L1}( d_0-d_{gt} ) + \sum_{i=1}^{n} \gamma^{n-i} \Vert d_i - d_{gt} \Vert _{1}
	\label{losst}
\end{equation}
where $\mathit{Smooth}_{L1}$ is defined by \cite{girshick2015fast}, $\gamma = 0.9$, $d_{gt}$ is ground truth disparity.
\begin{table*}[]
	\centering
	\begin{tabular}{c|ccccc|c}
		\toprule[1.5pt]
		Method & 
		Lac-GwcNet\cite{Lac2022Liu} & UPFNet \cite{upfnet} & ACVNet \cite{ACVNet} &DLNR \cite{2023DLNR} &IGEV-Stereo  \cite{igev2023xu} & MoCha-Stereo (Ours) \\ \midrule
		EPE (px)$\downarrow$
		& 0.75 
		& 0.71 
		& 0.48
		& 0.48 
		& \underline{0.47} 
		& \textbf{0.41} $_{\color[rgb]{1,0,0}(-12.77\%)}$  \\
		Time (s)$\downarrow$
		& 0.65 
		& \textbf{0.27} 
		& 0.48
		& \underline{0.30} 
		& 0.37 
		& 0.34  \\
		\bottomrule[1.5pt]
	\end{tabular}
	\caption{Quantitative evaluation on Scene Flow test set. The \textbf{best} result is bolded, and the \underline{second-best} result is underscored. The variations in the performance of our method compared to the optimal results of other methods are indicated in red font.}
	\label{SF}
	\vspace{-10pt}
\end{table*}
\begin{table*}[h]
	\centering
	\begin{tabular}{lcccccccc}
		\toprule[1.5pt]
		& \multicolumn{4}{c}{All} & \multicolumn{4}{c}{Reflective} \\
		\cmidrule(lr){2-5} \cmidrule(lr){6-9}
		& \textbf{Out-Noc} & Out-All & Avg-Noc & Avg-All & \textbf{Out-Noc} & Out-All & Avg-Noc & Avg-All \\
		\multicolumn{1}{l}{\multirow{-3}{*}{Method}} & \textbf{(\%)$\downarrow$} & (\%)$\downarrow$ & (px)$\downarrow$ & (px)$\downarrow$ & \textbf{(\%)$\downarrow$} & (\%)$\downarrow$ & (px)$\downarrow$ & (px)$\downarrow$ \\
		\hline
		GwcNet\cite{guo2019gwc}
		& {1.32 } & {1.70 } & {0.5 } & {0.5 } & {7.80 } & {9.28 } & {1.3 } & {1.4 } \\
		AcfNet\cite{zhang2020adaptive}
		& {1.17 } & {1.54 } & {0.5 } & {0.5 } & {6.93 } & {8.52 } & {1.8 } & {1.9 } \\		
		RAFT-Stereo \cite{lipson2021raft} &  {1.30 }   &  {1.66 }   &  \textbf{0.4 }  &  {0.5 }   &  {5.40 }   &  {6.48 }   &  {1.3 }   &  {1.3 }   \\
		HITNet\cite{tankovich2021hitnet}
		& {1.41 } & {1.89 } & \textbf{0.4 } & {0.5 } & {5.91 } & {7.54 } & {\underline{1.0} } & {1.2 } \\		
		CREStereo\cite{li2022practical}
		& {1.14 } & {1.46 } & \textbf{0.4 } & {0.5 } & {6.27 } & {7.27 } & {1.4 } & {1.4 } \\
		Lac-GwcNet \cite{Lac2022Liu} &  {1.13 }   &  {1.49 }   &  {0.5 }   &  {0.5 }   &  {6.26 }   &  {8.02 }   &  {1.5 }   &  {1.7 }   \\
		IGEV-Stereo \cite{igev2023xu}
		& {\underline{1.12} } & {\underline{1.44} } & \textbf{0.4 } & \textbf{0.4 } & {\underline{4.35} } & {\underline{5.00} } & {\underline{1.0} } & {\underline{1.1} } \\ 
		\hline
		MoCha-Stereo(Ours) & \textbf{1.06$_{\color[rgb]{1,0,0}(-5.36\%)}$} & \textbf{1.36 } & \textbf{0.4 } & \textbf{0.4 } & \textbf{3.83$_{\color[rgb]{1,0,0}(-11.95\%)}$ } & \textbf{4.50 } & \textbf{0.8 } & \textbf{0.9 }\\ 
		\bottomrule[1.5pt]
	\end{tabular}
	\caption{Results on the KITTI-2012 leaderboard. Out-Noc represents the percentage of erroneous pixels in non-occluded areas, Out-All denotes the percentage of erroneous pixels in the entire image. Avg-Noc refers to the end-point error in non-occluded areas, Avg-All indicates the average disparity error across the entire image. Error threshold is 3 px. }
	\label{2012}
	\vspace{-10pt}
\end{table*}
\section{Experiment}
\subsection{Implementation Details}
\label{4.1}
\quad MoCha-Stereo is implemented using the PyTorch framework, with the AdamW \cite{adamw} optimizer employed during training. 
For training and ablation experiments, our model was trained on the Scene Flow \cite{sceneflow} for 200k epochs, with a batch size of $8$, which is equipped with $2$ NVIDIA A6000 GPUs.
To evaluate the performance of our model, we conducted assessments using the  
KITTI-2012 \cite{kitti2012}, KITTI-2015 \cite{kitti2015}, Scene Flow \cite{sceneflow}, ETH3D \cite{eth3d}, and Middlebury \cite{middlebury}. The training and testing settings are consistent with \cite{lipson2021raft,igev2023xu}.
\begin{table}[]
\centering
\resizebox{1.0\linewidth}{!}{
\begin{tabular}{lcccccc}
	\toprule[2.5pt]
	& \multicolumn{3}{c}{All pixels (\%)$\downarrow$} & \multicolumn{3}{c}{Noc pixels (\%)$\downarrow$} \\
	\cmidrule(lr){2-4} \cmidrule(lr){5-7}
	\multicolumn{1}{l}{\multirow{-2}{*}{Method}} & bg & fg & \textbf{all} & bg & fg & \textbf{all} \\ \midrule
	GwcNet\cite{guo2019gwc}
	& {1.74 } & {3.93 } & {2.11 } & {1.61 } & {3.49 } & {1.92 } \\
	RAFT-Stereo\cite{lipson2021raft}
	& {1.58 } & {3.05 } & {1.82 } & {1.45 } & {2.94 } & {1.69 } \\
	CREStereo\cite{li2022practical}
	& {1.45 } & {2.86 } & {1.69 } & {1.33 } & {2.60 } & {1.54 } \\
	Lac-GwcNet\cite{Lac2022Liu}
	& {1.43 } & {3.44 } & {1.77 } & {1.30 } & {3.29 } & {1.63 } \\
	CFNet\cite{cfnet}& {1.54 } & {3.56 } & {1.81 } & {1.43 } & {3.25 } & {1.73 } \\		
	UPFNet\cite{upfnet}& {\underline{1.38} } & {2.85 } & {1.62 } & {1.26 } & {2.70 } & {1.50 } \\
	CroCo-Stereo\cite{croco2023}
	& {\underline{1.38} } & {\underline{2.65} } & {\underline{1.59} } & {1.30 } & {2.56 } & {1.51 } \\ 
	IGEV-Stereo\cite{igev2023xu}
	& {\underline{1.38} } & {2.67 } & {\underline{1.59} } & {\underline{1.27} } & {2.62 } & {\underline{1.49} } \\
	DLNR\cite{2023DLNR}
	& {1.60 } & {2.59 } & {1.76 } & {1.45 } & \textbf{2.39 } & {1.61 } \\ 
	\midrule
	MoCha-Stereo & \multirow{2}{*}{\textbf{1.36 }} & \multirow{2}{*}{\textbf{2.43 }} & \textbf{1.53 } & \multirow{2}{*}{\textbf{1.24 }} & \multirow{2}{*}{{\underline{2.42} }}  & \textbf{1.44 }
	\\ 
	(Ours) &  &  & \textbf{$_{\color[rgb]{1,0,0}-3.77\%}$ } & & & \textbf{$_{\color[rgb]{1,0,0}-3.36\%}$ }\\
	\bottomrule[2.0pt]
\end{tabular}}
\caption{Results on the KITTI-2015 leaderboard. Error threshold is 3 px. Background error is indicated by bg, and front-ground error by fg.}
\label{2015}
\vspace{-20pt}
\end{table}
\begin{figure*}[]
	\centering
	\includegraphics[width=\linewidth]{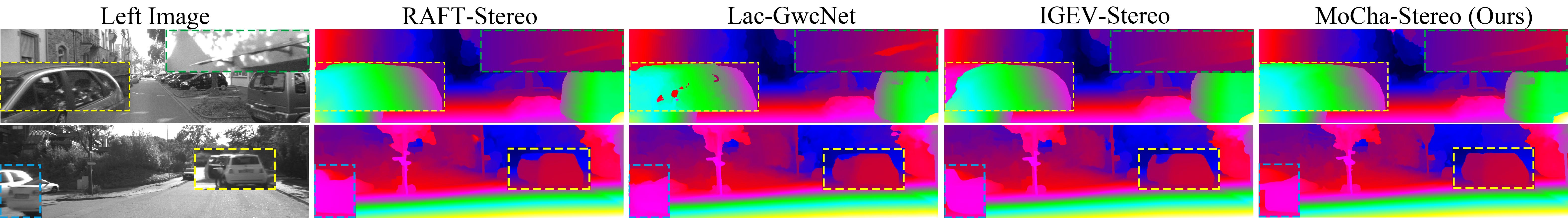}
	\caption{Visualisation on the KITTI dataset. We conducted comparisons with existing SOTA methods \cite{lipson2021raft,Lac2022Liu,igev2023xu}. 
	Our method accurately captures the edges of the left car and right roof in the first scene. In the second scene, it avoids confusion in the positions of the left two cars, and achieving a complete match of the edges of the right car doors. It is evident that our method excels in matching edge details.} \label{KITTI2015}
\end{figure*}
\begin{figure*}[]
	\centering
	\includegraphics[width=0.80\linewidth]{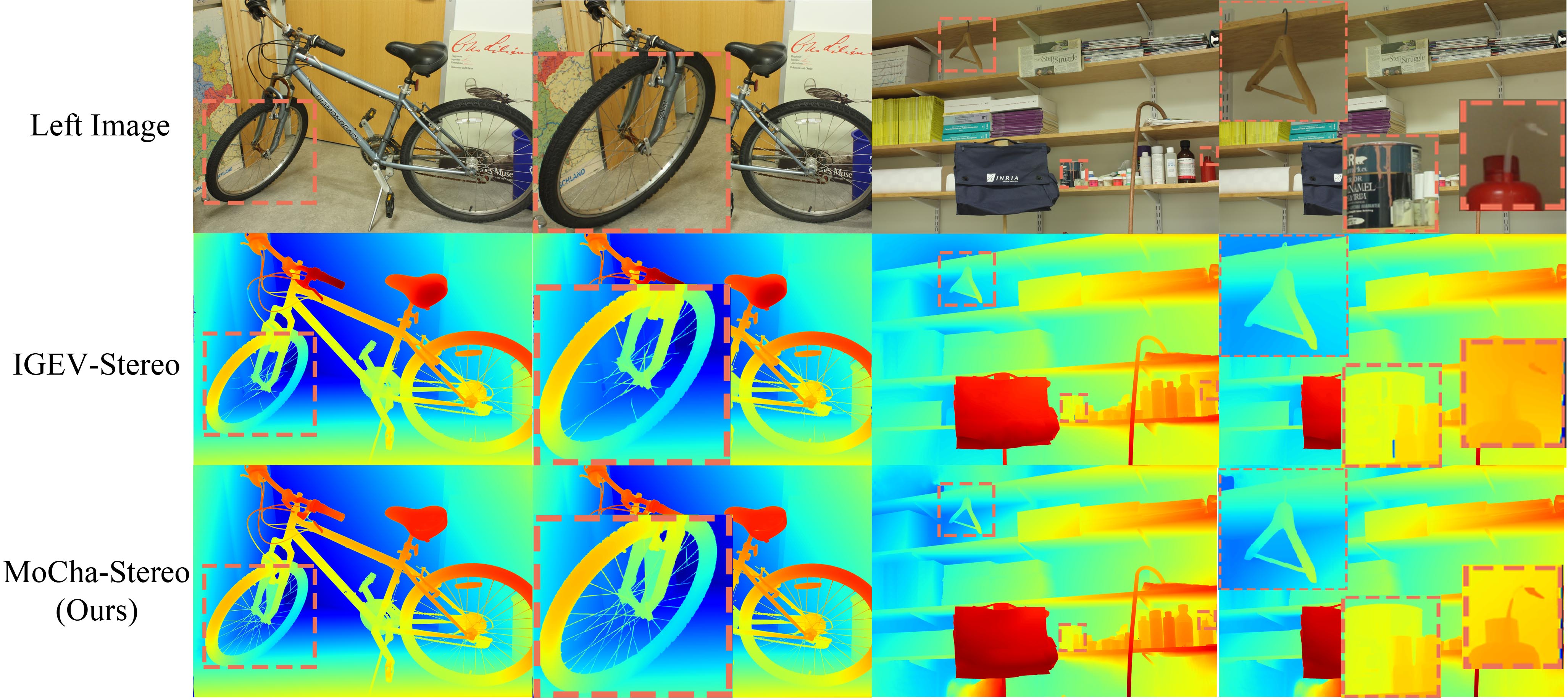}
	\caption{Visualisation on the Middlebury dataset. All results presented in this section demonstrate zero-shot generalization on the Scene Flow dataset. The odd-numbered columns show the original images, the even-numbered columns present zoomed-in details.} \label{Mid}
\end{figure*}
\subsection{Comparisons with State-of-the-art}
\label{4.2}
\quad We contrast the SOTA techniques on KITTI-2015 \cite{kitti2015}, KITTI-2012 \cite{kitti2012}, and Scene Flow \cite{sceneflow}. MoCha-Stereo achieves excellent performance on each of the aforementioned datasets. On the Scene Flow \cite{sceneflow} dataset, MoCha-Stereo achieves a new SOTA EPE of $0.41$, which surpasses IGEV-Stereo \cite{igev2023xu} by a margin of $\textbf{12.77\%}$. 
The quantitative comparisons are presented in Tab. \ref{SF}.

In order to validate the performance of MoCha-Stereo in real-world scenarios, we conducted experiments on the KITTI-2012 \cite{kitti2012} and KITTI-2015 \cite{kitti2015} benchmarks.  
MoCha-Stereo ranks $1st$ among all the methods submitted to these online benchmarks. Evaluation details are shown in Tab. \ref{2012} and Tab. \ref{2015}. 
We also provide visualizations of MoCha-Stereo and compare it with existing SOTA algorithms \cite{lipson2021raft,Lac2022Liu,igev2023xu} in Fig. \ref{KITTI2015}. 
Moreover, MoCha-Stereo achieves \textbf{$1st$} result in reflective regions, where determining geometric edges is often more challenging, as shown in Tab. \ref{2012}. 
MoCha-Stereo is the \textbf{first} algorithm to control Avg-Noc to within $0.8$ px under $5$ px error threshold and to control Out-Noc to less than $4\%$ for an error threshold of $3$ px among all published methods.
\subsection{Zero-shot Generalization}
\label{4.3}
\quad Due to the difficulty in obtaining a large amount of ground truth for real-world scenes, generalization ability is also crucial. We evaluate the generalization ability of MoCha-Stereo by testing it on the Middlebury \cite{middlebury} and ETH3D \cite{eth3d} datasets without fine-tune. 
As illustrated in Fig. \ref{Mid} and Tab. \ref{ZS}, our method exhibits SOTA performance in the zero-shot scenarios.
\begin{table}[]
	\centering
	\resizebox{1.0\linewidth}{!}{
		\begin{tabular}{lcccc}
			\toprule[1.5pt]
			& \multicolumn{3}{c}{Middlebury} & \multirow{2}{*}{ETH$\downarrow$} \\
			\cmidrule(lr){2-4}
			\multicolumn{1}{l}{\multirow{-2}{*}{Method}}& Full$\downarrow$ & Half$\downarrow$ & Quarter$\downarrow$&  \\ \hline
			PSMNet\cite{chang2018pyramid}&39.5&15.8&9.8&10.2\\
			GANet\cite{GANet}& 32.2 & 13.5 & 8.5&6.5 \\	
			DSMNet\cite{DSM}& 21.8 & 13.8 & 8.1&6.2 \\	
			CFNet\cite{cfnet} & 28.2 & 15.3 & 9.8&5.8 \\
			DLNR \cite{2023DLNR} & \underline{14.5} & 9.5 & 7.6 &23.1 \\
			IGEV-Stereo \cite{igev2023xu} & 15.2 & \underline{7.1} & \underline{6.2}&\underline{3.6} \\ \hline		
			MoCha-Stereo & \textbf{12.4} & \textbf{6.2} & \textbf{4.9}&\textbf{3.2} \\
			(Ours)&$_{\color[rgb]{1,0,0}-14.5\%}$&$_{\color[rgb]{1,0,0}-12.7\%}$
			&$_{\color[rgb]{1,0,0}-21.0\%}$
			&$_{\color[rgb]{1,0,0}-11.1\%}$ \\ 
			\bottomrule[1.5pt]
	\end{tabular}}
	\caption{Zero-shot evaluation on \cite{middlebury,eth3d}.  Every model undergoes scene flow training without fine-tuning on Middlebury and ETH3D dataset. The 2-pixel error rate is employed for Middlebury, and 1-pixel error rate for ETH3D.}
	\label{ZS}
	\vspace{-15pt}
\end{table}
\subsection{Extension to MVS}
\label{4.4}
\quad MoCha-Stereo has been extended as MoCha-MVS for application in the field of MVS.
Compared to recent learning-based MVS methods, MoCha-MVS achieves excellent performance by balancing accuracy and completeness. 
As shown in Tab. \ref{DTU}, our method outperforms SOTA methods specifically designed for MVS, indicating the excellent scalability of our approach.
\begin{table}[]
	\centering
	\begin{tabular}{lccc}
		\toprule[1.5pt]
		\multicolumn{1}{l}{Method} & \textbf{Ove.$\downarrow$} & Acc.$\downarrow$ & Comp.$\downarrow$ \\ \hline
		MVSNet\cite{yao2018mvsnet}
		& 0.462 & 0.396 & 0.527 \\
		CasMVSNet\cite{gu2020cascade}
		& 0.355 & \underline{0.325} & 0.385 \\
		PatchmatchNet\cite{wang2021patchmatchnet}
		&0.352&0.427&\textbf{0.277}\\
		IterMVS\cite{wang2022itermvs}
		& 0.363 & 0.373 & 0.354 \\
		CER-MVS\cite{ma2022multiview}
		& 0.332 & 0.359 & \underline{0.305} \\
		Vis-MVSNet\cite{zhang2023vis}
		& 0.365 & 0.369 & 0.361 \\		
		Miper-MVS\cite{zhou2023miper}& 0.345 & 0.364 & 0.327 \\
		DispMVS\cite{DispMVS}
		& \underline{0.339} & 0.354 & 0.324 \\ \hline
		MoCha-MVS & \textbf{0.319$_{\color[rgb]{1,0,0}(-5.90\%)}$} & \textbf{0.314} & 0.325 \\ 
		\bottomrule[1.5pt]
	\end{tabular}
	\caption{Quantitative evaluation on DTU \cite{jensen2014large} dataset expanded in MVS domain. Acc. means an indicator of accuracy, Comp. means an indicator of completeness, and Ove. means an indicator of the overall consideration of Acc. and Comp. (lower means better).}
	\label{DTU}
	\vspace{-15pt}
\end{table}

\subsection{Ablations}
\label{4.5}
\begin{figure*}[]
	\centering
	\includegraphics[width=\linewidth]{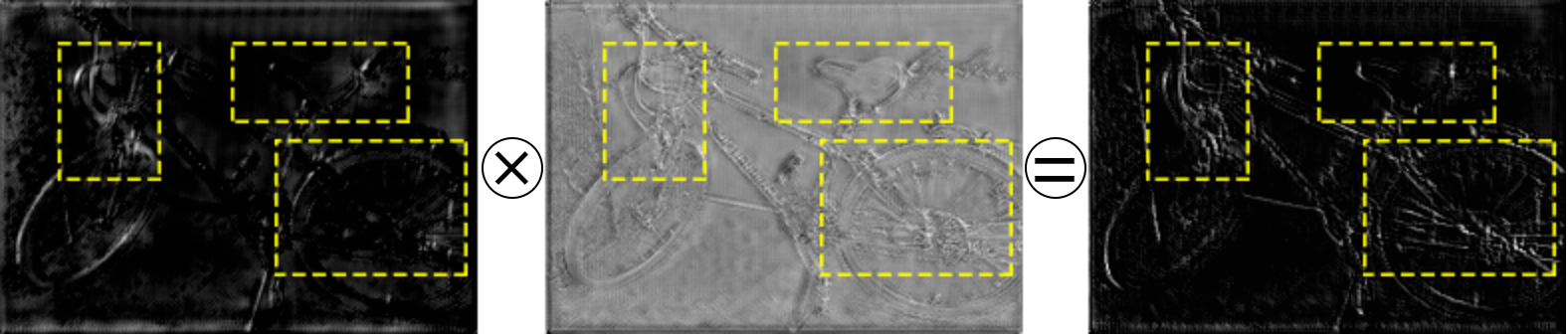}
	\caption{An example of one of the feature channels in visual form. The first picture shows the initial normal channel, and the last picture shows the visualization after paying attention to Motif Channels. The middle picture visualize a motif channel. It can be observed that the edge texture details are emphasized in the new feature channels.} \label{MCA}
\end{figure*}
\begin{table*}[]
	\centering
	\begin{tabular}{l|ccc|cc|cc}
		\toprule[1.5pt]
		\multirow{2}{*}{Model} & MCA for  & Correlation Volume  & \multirow{2}{*}{REMP} & \multirow{2}{*}{EPE (px)}  & D1-Error & \multirow{2}{*}{Time (s)} & \multirow{2}{*}{Params.(M)} \\
		& Feature Maps & guided by CAMP & &  & $\textgreater$3px(\%) & &     \\ \midrule 
		Baseline & & & & 0.458 & 2.536 & 0.33 & 18.31 \\ 
		+MCA&      \checkmark       &         &      & 0.449 & 2.492 & 0.33 & 18.32           \\
		+REMP&             &         &     \checkmark    &   0.445  & 2.469 & 0.33 & 20.70           \\
		+MCA+REMP&      \checkmark       &         &     \checkmark     & 0.438 & 2.434 & 0.33 & 20.71\\
		+MCCV&       \checkmark      &     \checkmark    &         & 0.423 & 2.358 & 0.34 & 18.35\\	
		Full model& \checkmark & \checkmark & \checkmark & \textbf{0.412} & \textbf{2.302} & 0.34 & 20.74 \\ \bottomrule[1.5pt]
	\end{tabular}
	\caption{Ablation study for MoCha-Stereo. The baseline employed in these experiments utilized EfficientNet \cite{tan2019efficientnet} as the backbone for IGEV-Stereo \cite{igev2023xu} with 16 iterations. The Time denotes the inference time on single NVIDIA A6000.}
	\label{abl}
	\vspace{-10pt}
\end{table*}
\begin{table}[]
	\centering
	\begin{tabular}{lcccccc}
		\toprule[1.5pt]
		\multirow{2}{*}{} & \multicolumn{6}{c}{Number of Iterations} \\ \cline{2-7}
		& 1     & 2    & 3    & 4    & 8  & 16  \\ \hline
		EPE (px) & 0.56  & 0.52 & 0.48 & 0.46 & 0.42 & \textbf{0.41} \\
		Time (s) & \textbf{0.19}  & 0.20 & 0.21 & 0.22 & 0.26 & 0.34 \\ 
		\bottomrule[1.5pt]
	\end{tabular}
	\caption{Ablation study for number of iterations.}
	\label{iter_num}
	\vspace{-20pt}
\end{table}
\quad To validate and comprehensively understand the architecture of our model, we conducted certain ablation experiments. Following \cite{lipson2021raft,igev2023xu}, all hyperparameter settings remained consistent with the pretraining phase for the Scene Flow dataset.\\
\textbf{Motif Channel Correlation Volume (MCCV).}
For all models in the ablation studies, we perform 16 iterations of updating at inference.
As shown in Tab. \ref{abl}, MCCV contributes to improved prediction accuracy. The decomposition of MCCV from coarse to fine stages into the actions on the feature map by MCA and on the Correlation Volume by CAMP results in a synergistic effect, leading to a reduction of 7.6\% in EPE (from 0.458 to 0.423). 
The effective improvement achieved by MCCV is attributed to its ability to address the bottleneck of existing feature extractors, which severely lose geometric edge information in some channels. This results in a more reasonable computation of the matching cost for geometric edges. As shown in Fig. \ref{MCA}, directing the attention of normal channels to motif channels summarizing repeated geometric textures in channel features enhances the clarity of edge textures in normal channels.\\
\textbf{Reconstruction Error Motif Penalty (REMP).}
As shown in Tab. \ref{abl}, REMP, by incorporating reconstruction error, learns motif information on the channels to understand high and low-frequency errors. The learned errors are then utilized as a penalty term to adjust the disparity map, resulting in a reduction of EPE by 2.8\% (from 0.458 to 0.445). This experimental result validates the effectiveness of REMP.\\
\textbf{Number of Iterations.}
MoCha-Stereo enhances the efficiency of iterations. 
As shown in Tab. \ref{iter_num}, MoCha-Stereo, with the information recovered by the Motif Channel, achieves SOTA results without the need for a large number of iterations. For instance, in a comparable inference time, we achieve a \textbf{40.8\%} reduction in EPE compared to UPFNet \cite{upfnet} (EPE 0.71 px, time 0.27 s) with \textbf{8} iterations. 
With only \textbf{4} iterations, MoCha-Stereo outperforms IGEV-Stereo \cite{igev2023xu} (EPE 0.47 px, time 0.37 s) by over 2.1\% in accuracy and saves \textbf{40.5\%} of the inference time. Information about \cite{upfnet,igev2023xu} can be obtained in Tab. \ref{SF}. 
Overall, MoCha-Stereo achieves SOTA performance even with a small number of iterations, allowing users to balance time efficiency and performance based on their specific needs.
\section{Conclusion and Future Work}

\quad We propose MoCha-Stereo, a novel stereo matching framework. MoCha-Stereo aims to alleviate edge mismatch caused by the geometric structure blurring of channel features. 
MCCV utilizes the geometric structure of repeated patterns in channel features to restore missing edge details and reconstructs the cost volume based on this novel channel feature structure. 
REMP penalizes the generation of the full-resolution disparity map based on the high and low-frequency information of the potential motif channel in the reconstruction error.
MoCha-Stereo showcases robust cross-dataset generalization capabilities. It ranks \textbf{1st} on the KITTI-2015 and KITTI-2012 Reflective online benchmarks and demonstrates SOTA performance on ETH3D, Middlebury, Scene Flow datasets and MVS domain. 
In the future, we plan to extend the motif channel attention mechanism to more processes in stereo matching, further enhancing the capability of algorithm for edge matching.
\\
\textbf{Acknowledgement.} 
This research is supported by Science and Technology Planning Project of Guizhou Province, Department of Science and Technology of Guizhou Province, China (Project No. [2023]159). Natural Science Research Project of Guizhou Provincial Department of Education, China (QianJiaoJi[2022]029, QianJiaoHeKY[2021]022).
{
    \small
    \bibliographystyle{ieeenat_fullname}
    \bibliography{main}
}


\end{document}